\documentclass[letterpaper]{article} 
\usepackage[]{aaai25}  
\usepackage{times}  
\usepackage{helvet}  
\usepackage{courier}  
\usepackage[hyphens]{url}  
\usepackage{graphicx} 
\usepackage{natbib}  
\usepackage{caption} 
\usepackage{algorithm}
\usepackage{algorithmic}
\usepackage{kotex}

\usepackage{newfloat}
\usepackage{listings}
\DeclareCaptionStyle{ruled}{labelfont=normalfont,labelsep=colon,strut=off} 
\floatstyle{ruled}
\newfloat{listing}{tb}{lst}{}
\floatname{listing}{Listing}
\usepackage{microtype}
\usepackage{booktabs}
\usepackage{amsmath}
\usepackage{amssymb}
\usepackage{mathtools}
\usepackage{amsthm}

\usepackage{comment}
\usepackage{adjustbox}
\usepackage{multirow}
\usepackage{subfigure}
\usepackage{xcolor}
\usepackage{bm}

\newcommand{\oursol}{\text{ICPAD}}
\newcommand{\domain}{d}
\newcommand{\skill}{z}
\newcommand{\DomainEnc}{\Psi_\text{E}}
\newcommand{\SkillEnc}{\Phi_\text{E}}
\newcommand{\SkillDec}{\Phi_\text{A}}
\newcommand{\SkillPrior}{\Phi_\text{R}}
\newcommand{\Retriever}{\Psi_\text{R}}
\newcommand{\DynamicPrompt}{\Psi_\text{D}}

\newcommand{\DataSet}{\mathcal{D}}

\newcommand{\StateSeq}{\bm{s}}
\newcommand{\ActionSeq}{\bm{a}}
\newcommand{\SubTraj}{\tau}

\DeclareMathOperator*{\expectation}{\mathbb{E}}

\title{In-Context Policy Adaptation via Cross-Domain Skill Diffusion}
\author{
    Minjong Yoo, Woo Kyung Kim, Honguk Woo\thanks{Corresponding Author}
}
\affiliations{
    Department of Computer Science and Engineering, Sungkyunkwan University\\
    \{mjyoo2, kwk2696, hwoo\}@skku.edu
}

\usepackage{bibentry}

\begin{document}

\maketitle

\begin{abstract}
In this work, we present an in-context policy adaptation (ICPAD) framework designed for long-horizon multi-task environments, exploring diffusion-based skill learning techniques in cross-domain settings. 
The framework enables rapid adaptation of skill-based reinforcement learning policies to diverse target domains, especially under stringent constraints on no model updates and only limited target domain data.
Specifically, the framework employs a cross-domain skill diffusion scheme, where domain-agnostic prototype skills and a domain-grounded skill adapter are learned jointly and effectively from an offline dataset through cross-domain consistent  diffusion processes. The prototype skills act as primitives for common behavior representations of long-horizon policies, serving as a lingua franca to bridge different domains. 
Furthermore, to enhance the in-context adaptation performance, we develop a dynamic domain prompting scheme that guides the diffusion-based skill adapter toward better alignment with the target domain.  
Through experiments with robotic manipulation in Metaworld and autonomous driving in CARLA, we show that our $\oursol$ framework achieves superior policy adaptation performance under limited target domain data conditions for various cross-domain configurations including differences in environment dynamics, agent embodiment, and task horizon. 
\end{abstract}

%

\section{Introduction}
Reinforcement learning (RL) has exhibited remarkable success in addressing sequential decision-making problems. However, a fundamental challenge emerges in the context of policy adaptation, the process of applying a policy trained in a source domain via RL to a different target domain, particularly for complex, long-horizon tasks~\cite{zhu2023transfer,zhao2020sim,hua2021learning, DCMRL, FIST}. This challenge escalates under practical restrictions that data availability is limited and direct interaction with the target domain is prohibited.

\begin{figure}[t]
    \centering
        \includegraphics[width=0.47\textwidth]{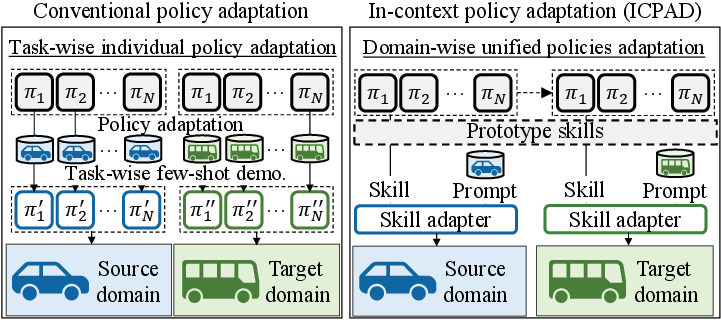}
    \caption{A middle-tier strategy for cross-domain policy adaptation in $\oursol$: as shown on the right, in $\oursol$, policies $\pi$ learned in the source domain with domain-agnostic prototype skills (in the middle-tier) are capable of adapting to the target domain in an in-context unified way by exploiting few-shot target domain data. 
    Unlike the conventional policy adaptation relying on \textit{policy-wise} model updates (shown on the left), $\oursol$ allows for simultaneous adaptation of multiple policies to the target domain through \textit{domain-wise} middle-tier adaptation.}
\label{fig:intro}
\end{figure}

To address this challenge, we explore skill-based RL approaches~\cite{SPIRL,SIMPL} in cross-domain environments, with the aim of facilitating the in-context adaptation of skill-based policies to different target domains under limited target data. We present an in-context policy adaptation framework, namely $\oursol$.
As illustrated in Figure~\ref{fig:intro}, $\oursol$ facilitates policy adaptation to target domains in a unified way, by learning transferable prototype skills and skill adapters offline, and implementing a middle-tier adaptation strategy.
Unlike conventional policy adaptation approaches, which typically require task-specific individual policy transfers, $\oursol$ offers a notable advantage in cross-domain multi-task environments. Its effectiveness stems from prototype skills that function as the middle-tier adaptation layer, linking policies and domains. These skills, which support intermediate representations or functional primitives of policies, facilitate their translation into domain-specific actions through dedicated skill adapters. 
In doing so, we adapt skill-based RL approaches to the context of cross-domain few-shot policy adaptation. That is, source domains have abundant data, while target domains are restricted to minimal data. 

In the offline learning phase of $\oursol$, we first establish the middle-tier adaptation layer, prototype skills, which encapsulate expert behavior patterns and are transferable across domains. 
Specifically, we develop a cross-domain skill diffusion scheme, incorporating the distribution matching-based cross-domain consistency into the skill diffusion process. This scheme allows for the domain alignment of action sequences, which are generated specifically for the target domain from the prototype skills. 
Then, using these prototype skills, which are established domain-agnostically from the offline dataset, RL policies are trained in the source domain.

%
In the in-context adaptation phase of $\oursol$, where the policies can be rapidly adapted to the target domain without model updates, we use a dynamic domain prompting scheme. It provides robust online guidance on the skill diffusion process toward enhanced alignment with the target domain, using only a few target data for in-context adaptation. 
These learning phases of $\oursol$ enable the in-context adaptation capabilities of skill-based policies.   

Through a series of experiments encompassing robotic manipulation in Metaworld~\cite{yu2020meta} and autonomous driving in CARLA~\cite{carla}, we demonstrate that $\oursol$ is applicable for a variety of cross-domain configurations including varied environment dynamics and different embodiment conditions. Furthermore, the policies learned via the framework ($\oursol$ policies) consistently outperform other state-of-the-art skill-based meta-RL methods. For example, in autonomous driving scenarios, $\oursol$ policies exhibit superior performance, achieving $11.6\%$ to $21.6\%$ higher normalized returns than the most competitive baseline DCMRL~\cite{DCMRL}.

Our main contributions are summarized as follows.
\begin{itemize}
    \item We present a novel in-context policy adaptation $\oursol$ framework to facilitate domain-wise rapid adaptation of skill-based policies across diverse domains without the need for online model updates. 
    \item We develop the cross-domain skill diffusion scheme, designed to integrate cross-domain consistency learning and skill diffusion, thereby achieving domain-agnostic prototype skills robustly and enabling their effective translation to domain-specific actions.  
    \item We adopt the dynamic domain prompting scheme, designed to enhance the adaptability of skill-based policies to the target domain.    
    \item Through extensive experiments with Metaworld and CARLA, we demonstrate $\oursol$'s effectiveness in a variety of cross-domain settings, establishing its superiority over several state-of-the-art methods.
\end{itemize}

\section{Preliminary}

\noindent \textbf{Skill-based RL.} 
With the goal of accelerating policy learning in complex environments, skill-based RL approaches leverage offline imitation learning on task-agnostic expert datasets, establishing a set of shared skills that capture expert behavior patterns. As implemented in~\cite{SPIRL}, these approaches encompass several models; a skill encoder maps expert sub-trajectories to a skill latent space and a skill decoder converts skill embeddings to action sequences. 
Unlike existing skill-based RL approaches that require online policy learning within target domains, our focus is on in-context policy adaptation to given target domains, prohibiting online environment interactions.

\noindent \textbf{Diffusion model.}
The diffusion model~\cite{DDPM} frames the generation process as an iterative denoising procedure, where a noisy input $x^K$ sampled from a standard normal distribution $p(x^K)$ is gradually denoised to recover the noiseless data $x^0$. That is
\begin{equation}
    p(x^0) = p(x^K) \prod_{k=1}^{K} p(x^{k-1}|x^k),
\end{equation}
where $k$ is the denoising timestep. 
As training a diffusion model $\epsilon$ involves optimizing a variational lower bound of $p(x^0)$, the below surrogate loss can be used.
\begin{equation}
\mathbb{E}_{k\sim[1,K],\eta\sim\mathcal{N}(0,I)}\left[||\eta - \epsilon(x_k,k)||_2^2\right]
\end{equation}
The noisy input $x^k$ is generated by adding the Gaussian noise to the original data $x^0$ as
\begin{equation}
  x^k = \sqrt{\bar{\alpha}^k} x^0 + \sqrt{1-\bar{\alpha}^k} \eta
  \label{equ:forward_process}
\end{equation}
where $\bar{\alpha}^k = \prod_{s=1}^K\alpha^k$ is a variance schedule.
This process can be viewed as estimating the noise $\eta$ that is added to the original data $x^0$ to generate the noisy input $x^k$. 
For generation, the diffusion model uses the iterative denoising chain such as
\begin{equation}
    x^{k-1} = \frac{1}{\sqrt{\alpha^k}} \left( x^{k} - \frac{1 - \alpha^k}{\sqrt{1-\bar{\alpha}^k}} \epsilon(x^k,k)\right).
  \label{equ:backward_process}
\end{equation}

\subsection{Problem Formulation}
\noindent \textbf{Cross-domain policy adaptation without model updates.} 
We consider the challenge of efficiently adapting RL policies to diverse target domains in long-horizon multi-task environment with limited target data. 
Specifically, we explore skill-based RL approaches utilizing a task-agnostic expert dataset collected from multiple domains. The dataset enables the creation of shareable, pre-trained skills, where each skill captures a reusable action sequence or behavior pattern. 

Our goal is to facilitate in-context adaptation of policies learned on these pre-trained skills, allowing for rapid policy deployment in the target domain without the need for gradient-based model updates or intensive environment interactions. This adaptation is conducted under limited target data conditions, leveraging only few-shot target domain data (i.e., no more than 5 trajectories across multiple tasks). 
The data requirements for offline skill learning are consistent with those of existing skill-based RL~\cite{SPIRL,SKIMO}. For the target domain, the data requirements align with the few-shot adaptation research~\cite{xu2022prompting, FIST, DCMRL}.
\section{Our Approach}

\subsection{Overall Framework}
\begin{figure}[t]
    \centering
    \includegraphics[width=0.47\textwidth]{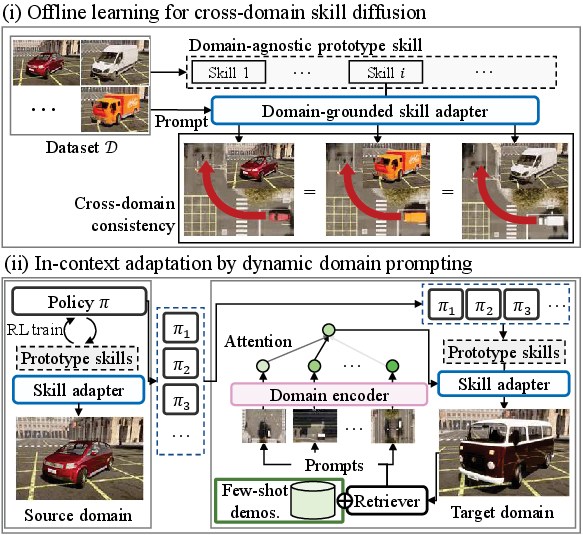}   
    \caption{Offline learning for cross-domain skill diffusion and in-context adaptation by dynamic domain prompting in $\oursol$: (i) In the offline learning phase, domain-agnostic prototype skills are learned jointly with a domain-grounded skill adapter that is prompted to generate consistent action sequences across domains. (ii) In the in-context adaptation phase, policies $\pi$ learned from the source domain are adapted to unseen target domains, facilitated by the skill adapter being prompted through retrieval-based attention.
}
\label{fig:procedure}
\end{figure}

To achieve in-context policy adaptation across domains, we present the $\oursol$ framework involving two phases: (i) offline learning for cross-domain skill diffusion and (ii) in-context adaptation by dynamic domain prompting. (i) In the offline learning phase, we first establish domain-agnostic prototype skills and a domain-grounded skill adapter by leveraging an offline dataset collected from multiple domains. (ii) In the latter phase of in-context adaptation, we facilitate the adaptation of policies learned with the prototype skills, by incorporating appropriate domain information into the context of diffusion-based skill translation, ensuring better alignment with the target domain. This dynamic domain prompting enables effective online policy adaptation, allowing for rapid deployment in diverse target domains.

Figure~\ref{fig:procedure} illustrates the concept of $\oursol$ framework. 
In the offline learning, we ensure that the distribution of prototype skills remains consistent across the domains, enabling the skill adapter to generate action sequences that accurately reflect the characteristics of both the skills and domains.
We also train the domain encoder through contrastive learning, which is used to dynamically retrieve proper prompts and applies attention to their embedding for in-context adaptation. 
Algorithm~\ref{algo:pretrain} lists the overall procedure of $\oursol$.

\begin{algorithm}[t]
    \begin{algorithmic}[1]
        \STATE{// Offline learning for cross-domain skill diffusion}
        \STATE{Initialize skill encoder $\SkillEnc$, adapter $\SkillDec$, prior $\SkillPrior$, domain encoder $\DomainEnc$,  Dataset $\mathcal{D}$}
        \WHILE{not converge}
           \STATE{Sample batches $\mathcal{B}, \mathcal{B}', \mathcal{B}_\text{C} \sim \mathcal{D}$}
           \STATE{Calculate the loss $\mathcal{L}_{\text{skill}}(\SkillEnc, \SkillDec)$ in~\eqref{equ:skill} using $\mathcal{B}$}
           \STATE{Calculate the loss $\mathcal{L}_{\text{cross-E}}(\SkillEnc, \SkillPrior)$ in~\eqref{equ:reg} using $\mathcal{B}$}
           \STATE{Calculate the loss $\mathcal{L}_{\text{cross-A}}(\SkillDec)$ in~\eqref{equ:cycle} using $\mathcal{B}'$}
           \STATE{Calculate the loss $\mathcal{L}_{\text{con}}(\DomainEnc)$ in~\eqref{equ:con} using $\mathcal{B}_\text{C}$}
           \STATE{Update $\SkillEnc, \SkillDec, \SkillPrior, \DomainEnc$ based on the sum of the losses}
        \ENDWHILE

        \STATE{// Iteration for each task (Line 12-24)}
        \STATE{// In-context adaptation by dynamic domain prompting}
        \STATE{Initialize skill-based policy $\pi$, dynamic prompting function $\DynamicPrompt$ by $\DomainEnc$, source env. $env$, replay buffer $\mathcal{D}_{R}$}
        \WHILE{not converge}
            \STATE{$s = env.\text{reset}()$}
            \LOOP
                \STATE{$z \sim \pi(\cdot|s)$, $d = \DynamicPrompt(\mathcal{D}_\text{R}, \StateSeq^h)$ in~\eqref{equ:ddp}}
                \STATE{$\ActionSeq = (a_0, ..., a_{H}) =\SkillDec(s, d, z)$}
                \STATE{Execute $\ActionSeq$ in $env$ and get $s$, add transitions in $\mathcal{D}_\text{R}$}
                \STATE{Sample batch $\mathcal{B}_\text{R} \sim \mathcal{D}_{R}$}
                \STATE{Optimize $\pi$ by~\eqref{equ:policy} using $\mathcal{B}_\text{R}$}
            \ENDLOOP
        \ENDWHILE
    \STATE{In-context adaptation with $\pi, \DynamicPrompt, \SkillDec$ in~\eqref{equ:infer} for target}
    \end{algorithmic}
    \caption{Learning procedure for ICPAD framework} 
    \label{algo:pretrain}
\end{algorithm}

\begin{figure*}[t]
    \centering
        \includegraphics[width=0.85\textwidth]{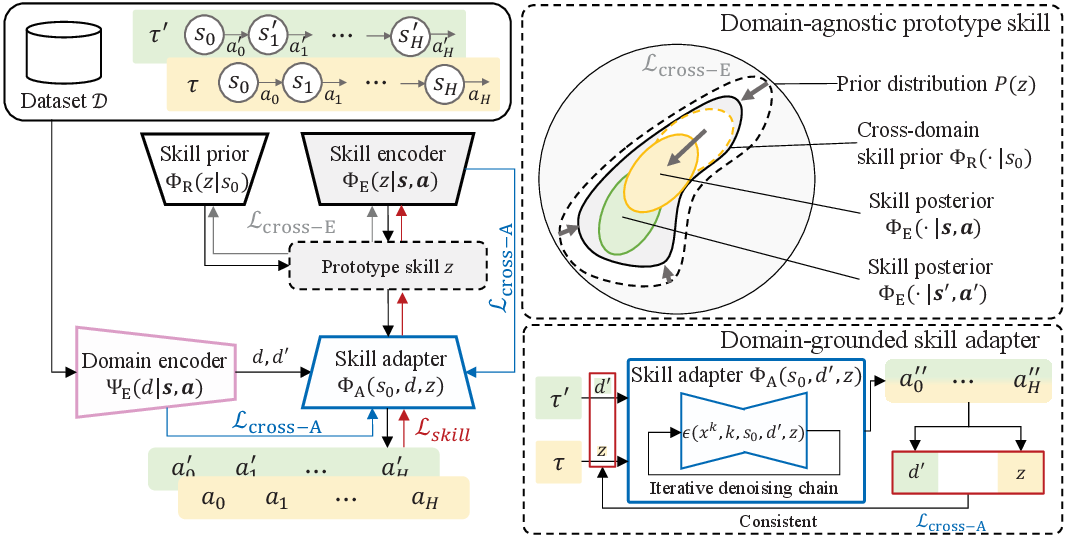}
    \caption{Cross-domain skill diffusion: we train the skill encoder $\SkillEnc$ for domain-agnostic prototype skills and the domain-grounded skill adapters $\SkillDec$ for domain-specific skill translation across domains. 
    These skill models are optimized using the skill imitation loss $\mathcal{L}_\text{skill}$ (Eq.~\eqref{equ:skill}), the cross-domain skill prior consistency loss $\mathcal{L}_\text{cross-E}$ (Eq.~\eqref{equ:reg}) for the skill encoder $\SkillEnc$ and skill prior $\SkillPrior$, as well as the cross-domain action consistency loss $\mathcal{L}_\text{cross-A}$ (Eq.~\eqref{equ:cycle}) for the skill adapter $\SkillDec$.}
    \label{fig:crost}
\end{figure*}

\subsection{Cross-domain Skill Diffusion}
\noindent \textbf{Domain-agnostic prototype skills.} 
To learn the prototype skills, we adapt the skill extraction approach~\cite{SIMPL} using the diffusion models with domain embedding. The skill encoder $\SkillEnc$ takes an $H$-step sub-trajectory $\SubTraj = (\StateSeq, \ActionSeq)$ and returns the distribution of prototype skill embedding $p(\skill|\StateSeq,\ActionSeq)$ where $\skill \in \mathcal{Z}$ is the skill embedding. The domain encoder $\DomainEnc$ takes an $H$-step sub-trajectory to return the distribution of domain embedding $p(\domain|\StateSeq, \ActionSeq)$. 
\begin{equation}
    \SkillEnc: (\StateSeq, \ActionSeq) \mapsto p(\skill|\StateSeq, \ActionSeq),\ \DomainEnc: (\StateSeq, \ActionSeq) \mapsto p(\domain| \StateSeq, \ActionSeq)
\end{equation}
Here, $\StateSeq$ and $\ActionSeq$ denote 
 a state sequence $(s_t, s_{t+1}, ... , s_{t+H})$ and an action sequence $(a_t, a_{t+1}, ..., a_{t+H})$, respectively. 
The skill adapter $\SkillDec$ produces an action sequence $\ActionSeq$ conditioned by state $s_t$, domain embedding $\domain$, and prototype skill embedding $\skill$. 
\begin{equation}
    \SkillDec: (s_t, \domain,\skill) \mapsto \ActionSeq
\end{equation}
For the skill adapter, we use diffusion models~\cite{DDPM}, in which an action sequence $\ActionSeq(=\!x^0)$ is generated from a random noisy input through an iterative denoising process. 
Accordingly, the skill adapter $\SkillDec$ implemented with a diffusion model $\epsilon$ takes additional inputs of a noised-corrupted action sequence $x^k$ and denoising step $k$.

To achieve the skill encoder and adapter, we implement the \textbf{skill imitation loss} integrating the diffusion loss and a regularize loss for the skill space, similar to~\cite{SPIRL}.
For a batch $\mathcal{B} = \{\tau\}$ sampled from $\mathcal{D}$, the skill imitation loss $\mathcal{L}_\text{skill}(\SkillEnc, \SkillDec)$ is defined as 
\begin{equation}
\begin{aligned}
 \expectation_{\substack{\skill \sim \SkillEnc, \domain \sim \DomainEnc, \mathcal{B}}} \Big[|| \eta- &\epsilon(x^{k}, k, s_t, \domain, \skill)||_2^2
\\&+ \beta D_\text{KL}\left(\SkillEnc(\cdot|\StateSeq,\ActionSeq), P(z)\right)\Big]
\end{aligned}
\label{equ:skill}
\end{equation}
where 
$k \sim [1, K]$ is a denoising step, $\eta\sim\mathcal{N}(0,I)$ is  noise sampled from the standard normal distribution, and $P(z)$ is the prior distribution for the prototype skill embedding. The regularization term $D_\text{KL}$ denotes the Kullback–Leibler divergence, and $\beta$ is a coefficient for regularization.

\noindent \textbf{Cross-domain consistency learning.}
To maintain cross-domain consistency for the skill encoder and ensure domain-agnostic characteristics in prototype skills, we use the cross-domain skill prior $\SkillPrior$. This model is responsible for inferring the prior distribution of skills based on states. The skill prior is formulated as
\begin{equation}
    \SkillPrior: s_t \mapsto p(\skill|s_t).
\end{equation}
For a batch $\mathcal{B}$ sampled from $\DataSet$, the \textbf{cross-domain skill prior consistency loss} $\mathcal{L}_\text{cross-E}(\SkillEnc, \SkillPrior)$ for the skill encoder and prior  is defined as 
\begin{equation}
    \begin{aligned}
     \mathbb{E}_{\SkillEnc, \SkillPrior, \mathcal{B}} \Big[&D_\text{KL}(\SkillEnc(\cdot|\StateSeq, \ActionSeq), sg(\SkillPrior (\cdot|s_t))) 
     \\&+ \mu D_\text{KL}(sg(\SkillEnc(\cdot|\StateSeq, \ActionSeq)), \SkillPrior (\cdot|s_t))\Big].
    \end{aligned}
    \label{equ:reg}
\end{equation}
Here, $sg$ is a stop gradient function, and $\mu$ is a hyperparameter.
By jointly training the skill encoder and the skill prior, the skill encoder is optimized to align with the distribution of skill prior relative to states, independent of domains. 

To maintain cross-domain consistency in the action sequences generated by the diffusion-based skill adapter, we align the prototype skill embedding $z$ and domain embedding $d$ extracted from the generated action sequences. 
For a batch $\mathcal{B}' = \{\tau, \tau' = (\StateSeq', \ActionSeq')\}$ sampled from $\mathcal{D}$, $z \sim \Phi_\text{E}(\mathbf{s}, \mathbf{a})$ and $d' \sim \Psi_\text{E}(\mathbf{s}', \mathbf{a}')$, \textbf{the cross-domain action consistency loss} $\mathcal{L}_\text{cross-A}(\SkillDec)$ for the skill adapter is defined as 
\begin{equation}
    \begin{aligned} 
    \expectation_{\substack{\skill \sim \SkillEnc, \domain' \sim \DomainEnc, \mathcal{B}'}} \Big[&D_\text{KL}(\DomainEnc(\cdot|\StateSeq, \SkillDec(s_t, \domain', \skill)), \DomainEnc(\cdot|\StateSeq',\ActionSeq')) \\
     +& D_\text{KL}(\SkillEnc(\cdot | \StateSeq, \SkillDec(s_t, \domain', \skill), \SkillEnc(\cdot| \StateSeq,\ActionSeq))\Big].
    \end{aligned}
    \label{equ:cycle}
\end{equation}
This ensures that the action sequences generated by the skill adapter are not only appropriate for the specific domain but also consistent across different domains. 
Figure~\ref{fig:crost} illustrates the cross-domain skill diffusion, of which the training involves domain-agnostic prototype skills and a domain-grounded skill adapter.
The joint learning approach with the loss formulation of Eq.~\eqref{equ:skill},~\eqref{equ:reg}, and~\eqref{equ:cycle} establishes the shareable prototype skill embedding across various domains, enabling the skill adapter to be effective for a wide range of domains and skills.
 
\begin{table*}[t]
    \footnotesize
    \centering
    \footnotesize
    \begin{tabular}{l ccc ccc}
        \toprule 
        Domain factor & \multicolumn{3}{c}{Action noise} & \multicolumn{3}{c}{Wind} \\
        \cmidrule(rl){1-1} \cmidrule(rl){2-4} \cmidrule(rl){5-7}
        Domain disparity & Low & Medium & High & Low & Medium & High \\
        \midrule
        \multicolumn{5}{l}{\textbf{Evaluation on Metaworld}} \\
        \midrule
        DiffBC+FT & $11.3 \pm 0.9\%$  & $8.2 \pm 0.5\%$ & $6.0 \pm 0.8\%$ & $11.7 \pm 0.3\%$  & $7.3 \pm 0.5\%$  & $4.8 \pm 0.3\%$ \\
        SPiRL & $18.5 \pm 1.3 \%$ & $15.1 \pm 1.1 \%$ & $9.7 \pm 0.6 \%$  & $12.8 \pm 1.1 \%$  & $8.9 \pm 1.3 \%$ & $6.1 \pm 1.2 \%$ \\
        SPiRL$^*$ & $35.7 \pm 4.1\%$ & $24.4 \pm 3.9\%$ & $23.1 \pm 3.4\%$ & $27.8 \pm 3.5\% $  & $22.1 \pm 4.1\%$ & $15.3 \pm 2.7\%$ \\
        FIST & $62.5 \pm 7.0\%$  & $51.3 \pm 6.9\%$  & $43.1 \pm 9.3\%$ & $63.1 \pm 4.6\%$ & $ 45.6 \pm 4.4\%$ & $ 33.7 \pm 5.9\%$ \\
        DCMRL & $ 77.3 \pm 6.2\%$ & $ 68.1 \pm 8.9\%$& $ 51.8 \pm 4.5\%$ & $ 71.0 \pm 4.8\%$ & $60.4 \pm 7.2\% $ & $46.1 \pm 8.0\% $ \\
        $\oursol$ (Ours) & \textbf{95.2} $\pm$ \textbf{4.9}\% & \textbf{85.4} $\pm$ \textbf{7.1}\% & \textbf{81.6} $\pm$ \textbf{7.4}\%& \textbf{89.1} $\pm$ \textbf{10.3}\% & \textbf{83.6} $\pm$ \textbf{7.5}\% & \textbf{74.9} $\pm$ \textbf{8.8}\% \\
        \midrule
        \multicolumn{5}{l}{\textbf{Evaluation on multi-stage Metaworld}} \\
        \midrule
        DiffBC+FT & $48.1 \pm 2.7\%$ & $45.3 \pm 2.5\%$   & $43.8 \pm 1.6\%$ & $38.2 \pm 2.2\%$ & $37.0 \pm 0.9\%$ & $36.2 \pm 2.1\%$ \\
        SPiRL & $ 52.5 \pm 0.8\% $& $50.6 \pm 2.3\%$ & $47.6 \pm 1.0\%$  & $51.1 \pm 0.9\%$ & $47.7 \pm 1.0\%$  & $46.3 \pm 2.2\%$ \\
        SPiRL$^*$ & $58.2 \pm 7.2\% $  & $56.1 \pm 6.5\%$   & $54.7 \pm 5.6\%$   & $57.0 \pm 4.9\% $& $54.3 \pm 5.1\% $ & $54.1 \pm 6.1\%$ \\
        FIST  & $89.4 \pm 1.3\%$   & $78.3 \pm 2.3\%$  & $68.9 \pm 3.0\%$    & $85.2 \pm 2.9\%$ & $75.5 \pm 1.8\%$& $64.3 \pm 2.0\%$\\
        DCMRL & $ \textbf{96.9} \pm \textbf{0.9}$\%  & $ 90.0 \pm 1.7\%$  & $ 81.7 \pm 2.7\%$   & $ 91.0 \pm 4.2\%$  & $ 81.9 \pm 2.6\%$  & $ 76.1 \pm 4.0\%$ \\
        $\oursol$ (Ours) & \textbf{96.9} $\pm$ \textbf{1.5}\%  & \textbf{94.8} $\pm$ \textbf{1.3}\%   & \textbf{90.2} $\pm$ \textbf{1.7}\% & \textbf{92.8} $\pm$ \textbf{1.5}\% & \textbf{90.6} $\pm$ \textbf{1.9}\%& \textbf{85.2} $\pm$ \textbf{1.8}\%\\
        
        \bottomrule
    \end{tabular}
    \caption{Adaptation performance in Metaworld and multi-stage Metaworld w.r.t. environment dynamics: 
    The performance in success rate is represented in 95\% confidence intervals with 5 seeds. For each domain (Action noise and Wind), we test 3 different domain disparity settings (Low, Medium, High) between the source, which includes offline data and online source domains, and the target domain.  
    }
    \label{tab:metaworld}
\end{table*}
\subsection{Dynamic Domain Prompting}\label{subsec:learningpolicy}

\noindent \textbf{Contrastive learning for domain encoder.} 
To train the domain encoder $\DomainEnc$, we adopt contrastive learning~\cite{schroff2015facenet}. We sample a batch $\mathcal{B}_\text{C} = \{(\SubTraj, \SubTraj^+), (\SubTraj, \SubTraj^-)\}$, where $(\SubTraj, \SubTraj^+)$ denotes a positive pair from the same domain, and $(\SubTraj, \SubTraj^-)$ denotes a negative pair. Then, for positive domain embedding $\domain^+\! \sim \DomainEnc(\cdot |\SubTraj^+\!=\!(\StateSeq^+, \ActionSeq^+))$ and negative domain embedding $\domain^-\! \sim \DomainEnc(\cdot |\SubTraj^-\!=\!(\StateSeq^-, \ActionSeq^-))$, the \textbf{contrastive loss} $\mathcal{L}_\text{con}(\DomainEnc)$ is defined as
\begin{equation}
 \mathbb{E}_{d \sim \DomainEnc, \mathcal{B}_\text{C}} \left[\max\{0, ||\domain - \domain^+|| - ||\domain - \domain^-|| + \delta\}\right] 
\label{equ:con}
\end{equation}
where $\delta$ is the margin of contrastive learning. 
Additionally, we employ the skill adapter to generate sub-trajectories from random domain embedding with the same skill embedding for negative samples.

\noindent \textbf{Retrieval-based prompt attention.}
The retriever $\Retriever$ samples $m$ prompts from target domain data (demonstrations) $\mathcal{T}$, based on a multinomial softmax distribution. 
The likelihood of retrieving a prompt $\hat{\tau}_i$ is determined by the sum of Euclidean distance $w_i$ between the state history $\StateSeq^{h} = (s_{t-H}, ...,s_{t})$ sampled from the policy and the states in $\mathcal{T}$.
\begin{equation}
    \hat{\mathbf{\tau}} = \{\hat{\tau}_1=(\hat{\StateSeq}_1, \hat{\ActionSeq}_1), \cdots, \hat{\tau}_m\} = \Retriever(\mathcal{T})
\end{equation} 
We also leverage the distance to incorporate attention into the domain embedding, ensuring that the embedding is properly chosen in the absence of data specific to the current task. Thus, during policy learning and adaptation, the below dynamic prompting function $\DynamicPrompt$ is used.
\begin{equation}
\DynamicPrompt(\mathcal{T}, \StateSeq^{h}) = \sum_{\hat{\tau}_i \sim \Phi_{R}(\mathcal{T})} \left[\frac{w_i^{-1}}{\sum^m_{i=1} w_i^{-1}}\DomainEnc(\hat{\StateSeq}_i, \hat{\ActionSeq}_i)\right]
\label{equ:ddp}
\end{equation}

\noindent \textbf{Policy learning and adaptation.} The skill-based policy $\pi$ is trained with the skill adapter $\SkillDec$ and the domain encoder $\DomainEnc$ in the source domain, and it is regularized by the cross-domain skill prior $\SkillPrior$.
Similar to~\cite{SPIRL}, for $\mathcal{B}_\text{R}$ sampled from replay buffer $\mathcal{D}_\text{R}$, the learning objective includes Q value maximization and $\SkillPrior$-regularization.
\begin{equation}
\begin{aligned}
     \max_\pi \mathbb{E}_{\mathcal{B}_\text{R}} [\bar{r}(s_t, z_t) - \lambda D_\text{KL}(\pi(\cdot|s_t), \SkillPrior (\cdot|s_t))]
\end{aligned}
\label{equ:policy}
\end{equation}
Here, $\bar{r}(s_t, z_t)$ is the sum of the rewards obtained by executing skill $z_t$ at state $s_t$ and $\lambda$ is a hyperparameter. In the source domain, we use the replay buffer for data  $\mathcal{T}$.

In policy adaptation, given few-shot target domain data $\mathcal{T}_\text{T}$, the policy $\pi$ (trained in the source domain) is capable of producing an action sequence by
\begin{equation}
    \ActionSeq = \SkillDec(s_t, \DynamicPrompt(\mathcal{T}_\text{T}, \StateSeq^{h}), z_t \sim \pi(\cdot|s_t))
    \label{equ:infer}
\end{equation}
for the target domain and state history $\StateSeq^{h} = (s_{t-H}, ... ,s_{t})$.

\section{Experiments}
We evaluate our framework in environments with diverse tasks and domains, including robotics manipulation in Metaworld~\cite{yu2020meta}, and autonomous driving with diverse routes in CALRA~\cite{carla}.

\noindent \textbf{Baselines.}
\textbf{DiffBC+FT}~\cite{DiffBC} is a diffusion-based imitation learning baseline that incorporates fine-tuning.  
\textbf{SPiRL}~\cite{SPIRL} is a skill-based RL baseline for long-horizon environments. We assess two variations of SPiRL for few-shot adaptation: the standard SPiRL which fine-tunes a skill-based policy and SPiRL$^*$ which additionally fine-tunes the skill decoder (adapter).
\textbf{FIST}~\cite{FIST} is a skill-based few-shot imitation method, enabling rapid adaptation through a semi-parametric approach to skill determination. 
\textbf{DCMRL}~\cite{DCMRL} is a state-of-the-art skill-based meta-RL method that enhances adaptability by decoupling the Gaussian context and the skill decoder to adjust to environmental contexts. 

\begin{table*}[t]
    \footnotesize
    \centering

    \begin{tabular}{l ccc ccc}
        \toprule
        Domain factor & \multicolumn{3}{c}{Vehicle embodiment} & \multicolumn{3}{c}{Vehicle embodiment + Weather} \\
        \cmidrule(rl){1-1} \cmidrule(rl){2-4} \cmidrule(rl){5-7}
        Domain disparity & Low & Medium & High & Low & Medium & High \\
        \midrule
        DiffBC+FT & $21.4 \pm 0.9\%$  & $17.6 \pm 0.7\%$ & $15.1 \pm 0.7\%$ & $19.7 \pm 0.8\%$ & $16.9 \pm 0.9\%$ & $13.2 \pm 1.3\%$ \\
        SPiRL  & $26.5 \pm 1.2\%$ & $22.0 \pm 1.2\%$ & $20.2 \pm 1.4\%$ & $24.3 \pm 1.0\%$ & $20.7 \pm 1.3\%$ & $17.2 \pm 1.5\%$\\
        SPiRL$^*$  & $33.1 \pm 1.8\%$ & $26.4 \pm 0.8\%$ & $23.8 \pm 2.1\%$ & $32.1 \pm 1.5\%$ & $25.6 \pm 1.4\%$ & $19.9 \pm 1.7\%$ \\
        FIST  & $60.8 \pm 2.6\%$ & $51.0 \pm 5.7\%$ & $47.7 \pm 1.6\%$ & $58.9 \pm 2.6\%$ & $49.5 \pm 1.9\%$ & $43.7 \pm 2.7\%$ \\
        DCMRL  & $68.4 \pm 2.2\%$ & $59.7 \pm 1.3\%$ & $56.6 \pm 2.7\%$ & $67.5 \pm 3.2\%$ & $58.5 \pm 3.3\%$ & $51.9 \pm 3.2\%$ \\
        $\oursol$ (Ours)  & \textbf{81.3} $\pm$ \textbf{2.8}\% & \textbf{77.3} $\pm$ \textbf{4.2}\% & \textbf{73.4} $\pm$ \textbf{2.0}\% & \textbf{79.1} $\pm$ \textbf{1.8}\% & \textbf{76.3} $\pm$ \textbf{2.7}\% & \textbf{73.5} $\pm$ \textbf{2.0}\% \\
        \bottomrule
    \end{tabular}
    \caption{Adaptation performance in CARLA w.r.t. embodiment:
    The performance is normalized based on the maximum return by an expert policy in the target domain.}
    \label{tab:carla}
\end{table*}
\subsection{Cross-domain Adaptation Performance}
\noindent \textbf{Cross-domain w.r.t. environment dynamics.} 
To evaluate the adaptation performance across domain changes with varied environmental dynamics, we use the Metaworld and multi-stage Metaworld with additional action noise and wind settings. The Metaworld includes the multiple directional reach task, and the multi-stage Metaworld provides long-horizon scenarios for training agents in complex sequential sub-tasks such as closing a drawer and pressing a button. 
For the multi-stage Metaworld, we assess performance by measuring the success rate for each sub-task, calculated as the average ratio of successfully completed sub-tasks.
For in-context policy adaptation, we use 5-shot demonstrations for each task. 
The domain disparity indicates the difference between the source, which includes offline data and online source domains, and target domains (i.e., differences in action noise or wind). 
Details of the domain disparity settings are in Appendix.

In Table~\ref{tab:metaworld}, $\oursol$ consistently outperforms the baselines, demonstrating its superiority with an average success rate that is 14.0\% higher than the closest competitor, DCMRL.
SPiRL shows low performance due to its skill decoder's limitations with offline data; SPiRL* improves by fine-tuning more components but lacks stability and efficiency. 
FIST and DCMRL, while adapting skill-based policies, suffer significant performance drops of 20.1\% and 22.6\% respectively due to insufficient skill generalization as domain disparity increases. In contrast, our $\oursol$ shows only a slight performance degradation of 10.5\%.
$\oursol$ effectively employs cross-domain consistent learning for domain-agnostic prototype skills, which generalize skills across domains, and the domain-grounded skill adapter transforms these prototype skills into action sequences tailored to the specific characteristics of each domain.
Using this approach, $\oursol$ consistently achieves better performance, even in diverse and unseen environmental dynamics.

\noindent \textbf{Cross-domain w.r.t. embodiment.} 
The CARLA simulator provides real-world scenarios for training agents in autonomous driving across diverse domains, including various vehicle embodiment and weather conditions. 
For multi-task scenarios, we configure different starting points and destinations to have various routes in navigation tasks.
Table~\ref{tab:carla} presents the normalized returns in CARLA. 
$\oursol$ consistently surpasses the most competitive baseline, DCMRL, by $11.6\%$ to $21.6\%$.
Unlike FIST and DCMRL, which require direct adaptation of skill-based policies, the skill prompting approach allows $\oursol$ to
simplify the complexity of the environment by enabling adaptation at the skill abstraction level.
This facilitates the effective use of skill-based policies across various domains, even in complex environments. 

\noindent \textbf{Cross-domain w.r.t. horizon length.} 
Table~\ref{tab:horizon} shows the adaptation performance across domain changes on task horizon, presenting the success rates in the multi-stage Metaworld.
All the skill-based baselines effectively address the changes in horizon length.
Among them, $\oursol$ shows slightly better performance than the most competitive baseline DCMRL, by a margin of $0.3\%$ to $1.7\%$. 
This robust performance can be attributed  to dynamic domain prompting that selects appropriate prompts for each state, enabling effective adaptation to domains with extended horizons.

\begin{table}[t]
    \footnotesize
    \centering
    \begin{adjustbox}{width=0.47\textwidth}
    \begin{tabular}{lccc}
        \toprule
        Domain factor & \multicolumn{3}{c}{Horizon length} \\
        \cmidrule(rl){1-1} \cmidrule(rl){2-4}
        Domain disparity & Low & Medium & High \\ 
        \midrule
        DiffBC+FT  & $87.7 \pm 2.8\%$ & $85.1 \pm 3.5\%$ & $77.6 \pm 4.7\%$ \\
        SPiRL  & $92.1 \pm 1.2\%$ & $92.5 \pm 1.1\%$  & $82.4 \pm 1.6\%$ \\
        SPiRL$^*$ & $97.8 \pm 1.7\% $& $94.5 \pm 2.1\% $ & $86.0 \pm 3.3\%$ \\
        FIST & \textbf{99.4} $\pm$ \textbf{0.5}\% & $95.2 \pm 1.1\%$& $94.2 \pm 1.0\%$\\
        DCMRL & $ 99.1 \pm 0.4\%$  & $96.3 \pm 0.5\%$  & $ 94.9 \pm 0.8\%$ \\
        $\oursol$ (Ours) & \textbf{99.4} $\pm$ \textbf{0.2}\%& \textbf{97.8} $\pm$ \textbf{0.6}\%& \textbf{96.6} $\pm$ \textbf{0.7}\%\\
        \bottomrule
    \end{tabular}
    \end{adjustbox}
    \caption{Adaptation performance in multi-stage Metaworld w.r.t. horizon length.}
    \vskip -0.1in
    \label{tab:horizon}
\end{table}

\subsection{Analysis}
\noindent \textbf{Limited data availability for tasks.}
Table~\ref{tab:dataset_analysis} shows the average success rate of $\oursol$ with respect to data availability for the tasks from the target domain in the multi-stage Metaworld.
In this test, we use only a selected percentage of tasks, each supported by 5-shot demonstrations. For tasks without data, we use data from the most similar task as a substitute.
With only $8\%$ of the tasks represented by demonstrations in the target domain, our $\oursol$ exhibits a slight performance decline of $6.8\%$, underscoring its robustness. In contrast, a $32.0\%$ decrease is observed with DCMRL under the same conditions. 
This demonstrates the efficiency of our unified domain-wise policy adaptation, which utilizes skill prompting even with strictly limited data from the target domain.
\begin{table}[h]
    \footnotesize
    \centering
    \begin{adjustbox}{width=0.47\textwidth}
    \begin{tabular}{lccccc}
    \toprule
    Data availability & 8\% & 20\% & 100\%\\
    \midrule
    FIST & $22.0 \pm 0.9\%$ & $47.2 \pm 2.2\%$ & $75.5 \pm 1.8\%$\\
    DCMRL & $49.9\pm 2.8\%$ & $70.3 \pm 1.7\%$ & $81.9 \pm 2.6\%$\\
    $\oursol$ & $\textbf{83.8} \pm \textbf{2.1}$\% & $\textbf{87.9} \pm \textbf{1.7}$\% &\textbf{90.6} $\pm$ \textbf{1.9}\% \\
    \hline
    \end{tabular}
    \end{adjustbox}
    \caption{Impact of the data availability for tasks.}
    \label{tab:dataset_analysis}
\end{table}

\noindent\textbf{Expansion of language-based prompts.}
In real-world scenarios, there is a growing demand for an interactive human-agent system, where agents follow language instructions.
To evaluate our framework in such a context, we utilize a dataset labeled with language instructions specific to each domain (e.g., The wind blows from the left).
We integrate these language instructions as prompts in $\oursol$, using an additional language encoder that is optimized by the domain encoder. 
For comparison, we implement L-DCMRL, which trains the language encoder with the context encoder.
In Table~\ref{tab:lan}, $\oursol$ maintains a superior performance compared to L-DCMRL.
This specifies the ability of $\oursol$ to effectively utilize prompts from different modalities, achieving a similar performance to that obtained with sub-trajectories.
\begin{table}[h]
    \footnotesize
    \centering
    \begin{adjustbox}{width=0.47\textwidth}
    \begin{tabular}{lccc}
        \toprule
        Domain factor & \multicolumn{3}{c}{Instructions related to wind} \\
        \cmidrule(rl){1-1} \cmidrule(rl){2-4}
        Domain disparity & Low & Medium & High \\ 
        \midrule
        L-DCMRL & $80.8 \pm 1.1\%$ & $75.4 \pm 1.5\%$ & $67.5 \pm 1.4\%$\\
        $\oursol$& \textbf{90.4} $\pm$ \textbf{2.1}\% & \textbf{87.1} $\pm$ \textbf{1.6}\%& \textbf{82.9} $\pm$ \textbf{1.4}\% \\
        \bottomrule
    \end{tabular}
    \end{adjustbox}
    \caption{Performance with language-based prompts in multi-stage Metaworld.}
    \label{tab:lan}
\end{table}

\begin{figure}[h]
    \centering
    \includegraphics[width=0.38\textwidth]{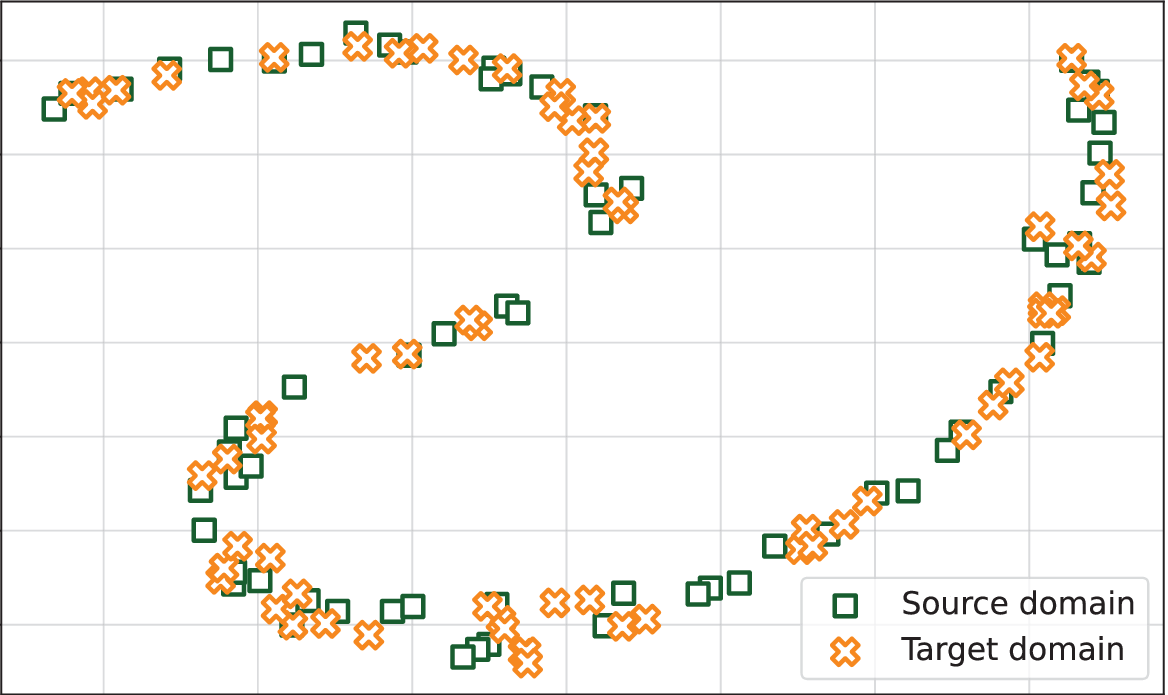}
    \label{fig:QA:b}
    \caption{Qualitative analysis for prototype skill embedding.
    }
    \label{fig:QA}
    \vskip -0.1in
\end{figure}

\noindent \textbf{Qualitative analysis for prototype skill embedding.}
Figure~\ref{fig:QA} visualizes the prototype skill embedding generated by $\oursol$ for multi-stage Metaworld tasks.
For different domains within the same task, the skill embeddings are closely paired together. This indicates that $\oursol$ effectively constructs shareable prototype skills, demonstrating that simply changing the prompt for the skill adapter allows seamless adaptation.
\subsection{Ablation Study}

We conduct ablation tests with the multi-stage Metaworld. 

\noindent\textbf{Ablation for cross-domain skill diffusion.} 
In Table~\ref{tab:abl1}, to show the effect of the cross-domain skill diffusion, we test two variants of $\oursol$; $\oursol$-A does not use the cross-domain action consistency loss for the skill adapter $\mathcal{L}_\text{cross-A}$ in~\eqref{equ:cycle}, and $\oursol$-D does not use diffusion models for the skill adapter.
As shown, $\oursol$ outperforms these two variants, achieving a performance gain of $5.4\%$ to $6.3\%$.
\begin{table}[h]
    \footnotesize
    \centering
    \begin{adjustbox}{width=0.46\textwidth}
    \begin{tabular}{lccc}
        \toprule
        Domain factor & \multicolumn{3}{c}{Wind} \\
        \cmidrule(rl){1-1} \cmidrule(rl){2-4}
        Domain disparity & Low & Medium & High \\ 
        \midrule
        $\oursol$-A & $79.6 \pm 2.2\%$& $76.4 \pm 3.0\%$& $53.9 \pm 2.1\%$\\ 
        $\oursol$-D & $87.4 \pm 3.3\%$ & $84.1 \pm 2.6\%$ & $78.9\pm 3.1\%$\\
        $\oursol$ & \textbf{92.8} $\pm$ \textbf{1.5}\% & \textbf{90.6} $\pm$ \textbf{1.9}\%& \textbf{85.2} $\pm$ \textbf{1.8}\%\\
        \bottomrule
    \end{tabular}
    \end{adjustbox}
    \caption{Ablation study for cross-domain skill diffusion.}
    \label{tab:abl1}
\end{table}

\noindent \textbf{Ablation for dynamic domain prompting.} 
In Table~\ref{tab:abl2}, to show the effect of the dynamic domain prompting, we test two variants of $\oursol$; $\oursol$-Fix does not use a retrieval approach, but it uses the average of entire data for domain embedding. $\oursol$-Avg does not use attention, but it uses the average of retrieved prompts for domain embedding.
The results show that $\oursol$ outperforms these variants, demonstrating an average performance gain of $1.6\%$ to $5.5\%$.
Although the performance gap is minimal at lower levels of domain disparity, it becomes more pronounced as the disparity increases. 
This is because retrieval-based attention supports flexible adaptation to unseen domains. 
\begin{table}[h]
    \footnotesize
    \centering
    \begin{adjustbox}{width=0.46\textwidth}
    \begin{tabular}{lccc}
        \toprule
        Domain factor & \multicolumn{3}{c}{Wind} \\
        \cmidrule(rl){1-1} \cmidrule(rl){2-4}
        Domain disparity & Low & Medium & High \\ 
        \midrule
        $\oursol$-Fix & $78.5 \pm 2.7$\%& $67.9 \pm 1.8$\%& $54.1 \pm 2.9$\%\\ 
        $\oursol$-Avg & $91.2 \pm 1.7$\%& $88.4 \pm 1.4$\%& $79.7 \pm 2.6$\%\\ 
        $\oursol$ & \textbf{92.8} $\pm$ \textbf{1.5}\% & \textbf{90.6} $\pm$ \textbf{1.9}\%& \textbf{85.2} $\pm$ \textbf{1.8}\%\\
        \bottomrule
    \end{tabular}
    \end{adjustbox}
    \caption{Ablation study for dynamic domain prompting.}
    \label{tab:abl2}
    \vskip -0.1in
\end{table}

\section{Related Works}
\textbf{Policy adaptation.}
In the application areas of autonomous systems where interaction with the target environment is limited or data from the target is scarce, policy adaptation methods have been investigated. 
Previous research primarily focused on searching a favorable initial weight~\cite{MAML, REPTILE}, learning policy-based task embeddings~\cite{VARIBAD}, or adjusting rewards to mitigate domain differences~\cite{DARC, DARA}.
However, these approaches often encounter challenges in more complex, data-restricted environments, particularly when online interaction with the target is not allowed or when only limited expert data is available. In contrast, our focus is on addressing these data limitations in the target domain by integrating skill diffusion with cross-domain consistency learning, thus enabling in-context policy adaptation across domains.

\noindent \textbf{Consistency learning.}
To achieve models with cross-domain capabilities, research is being conducted on extracting and transforming domain characteristics through the notion of consistency learning~\cite{CycleGAN, chen2019crdoco, wang2023cdac}.
Particularly, in the RL literature, to facilitate policy transfer, CDIO~\cite{CDIO} and RL-CycleGAN~\cite{RLCycleGAN} establish domain correspondence mappings via consistency learning.
Our cross-domain skill diffusion incorporates consistency learning within the diffusion-based skill translation, aiming for a unified policy adaptation strategy based on prototype skills at a middle-tier level.
Our work is the first to achieve in-context adaptation of skill-based policies, exploring the integration of skill diffusion and consistency learning.  

\noindent \textbf{Skill-based RL.}
To tackle complex tasks with challenging exploration, research efforts have been focused on incorporating short-horizon skills in hierarchical RL and multi-task RL~\cite{DIAYN,VIC,sharma2019dynamics}.
Recent works adopt offline skill imitation from task-agnostic expert datasets to accelerate policy learning~\cite{SPIRL,SIMPL,STAR}.
Our work also leverages skill imitation but distinguishes itself by concentrating on cross-domain skill learning and in-context policy adaptation. 

\section{Conclusion}
We presented the $\oursol$ framework to enable in-context policy adaptation across domains, effectively addressing the challenges posed by data limitations in the target domain.
Central to the framework are the domain-agnostic prototype skills and their domain-specific translation via the diffusion-based skill adapter, achieved through consistency learning.  
The framework also employs dynamic domain prompting to enhance the performance of in-context policy adaptation. 
In our future work, we plan to extend the framework to accommodate multi-modal datasets, aiming to explore the semantic interpretability and alignment across significantly different tasks and domains for embodied control applications. 

\section{Acknowledgements}
This work was supported by Institute of Information \& communications Technology Planning \& Evaluation (IITP) grant funded by the Korea government (MSIT),
(RS-2022-II220043 (2022-0-00043), Adaptive Personality for Intelligent Agents, 
RS-2022-II221045 (2022-0-01045), Self-directed multi-modal Intelligence for solving unknown, open domain problems,
No. Rs-2020-II201821, ICT Creative Consilience Program,
and RS-2019-II190421, Artificial Intelligence Graduate School Program (Sungkyunkwan University)),
the National Research Foundation of Korea (NRF) grant funded by the Korea government (MSIT)  (No. RS-2023-00213118),
IITP-ITRC (Information Technology Research Center) grant funded by the Korea government (MIST) (IITP-2024-RS-2024-00437633, 10\%),
and by Samsung Electronics.

\bibliography{ref}

\end{document}